\documentclass[11pt,a4paper]{article}
\PassOptionsToPackage{hyphens}{url}\usepackage{hyperref}
\usepackage[hyperref]{eacl2021}
\usepackage{times}
\usepackage{latexsym}
\usepackage{todonotes}
\usepackage{breakurl}

\usepackage{microtype}
\usepackage{booktabs} %
\usepackage[hang,flushmargin]{footmisc} %
\usepackage{comment}  %
\usepackage{amssymb}  %
\usepackage{caption}  %
\usepackage{subcaption}
\usepackage{amsmath}  %
\usepackage{xcolor}   %
\usepackage{graphicx} %
\usepackage{multirow}
\usepackage{mdframed}
\usepackage[capitalise]{cleveref}

\newmdenv[linecolor=white,backgroundcolor=white]{myframe}

\definecolor{pink}{RGB}{255, 204, 204}
\definecolor{purple}{RGB}{178, 102, 255}
\definecolor{lightred}{RGB}{226, 53, 53}
\definecolor{lightgreen}{RGB}{49, 144, 226}

\aclfinalcopy %

\title{ToxCCIn: Toxic Content Classification with Interpretability}

\author{ Tong Xiang $^1$ \quad Sean MacAvaney $^{1,2}$ \quad Eugene Yang $^1$ \quad Nazli Goharian $^1$  \\
$^1$ IR Lab, Georgetown University, USA
\quad $^2$ University of Glasgow, UK \\
\texttt{tx39@georgetown.edu}\quad \texttt{sean.macavaney@glasgow.ac.uk} \\ \texttt{eugene@ir.cs.georgetown.edu}\quad \texttt{nazli@ir.cs.georgetown.edu} \\
}

\date{}

\begin{document}
\maketitle
\begin{abstract}
Despite the recent successes of transformer-based models in terms of effectiveness on a variety of tasks, their decisions often remain opaque to humans. Explanations are particularly important for tasks like offensive language or toxicity detection on social media because a manual appeal process is often in place to dispute automatically flagged content. In this work, we propose a technique to improve the interpretability of these models, based on a simple and powerful assumption: a post is at least as toxic as its most toxic span. We incorporate this assumption into transformer models by scoring a post based on the maximum toxicity of its spans and augmenting the training process to identify correct spans. We find this approach effective and can produce explanations that exceed the quality of those provided by Logistic Regression analysis (often regarded as a highly-interpretable model), according to a human study.
\end{abstract}

\section{Introduction}
The rapidly increasing usage of social media has made communication easier but also has enabled users to spread questionable content~\citep{doi:10.1080/1369118X.2017.1293130}, which sometimes even leads to real-world crimes~\citep{johnson2019hidden,home2017hate,center2017year}. To prevent this type of speech from jeopardizing others' ability to express themselves in online communities, many platforms prohibit content that is considered abusive, hate speech, or more generally, toxic. To enforce such policies, some platforms employ automatic content moderation, which uses machine learning techniques to detect and flag violating content~\citep{10.1145/2740908.2742760,10.1145/2872427.2883062,macavaney:plosone2019-hate}.

Leveraging the development of pre-trained language models~\citep{devlin-etal-2019-bert,liu2019roberta} and domain transfer learning~\citep{gururangan-etal-2020-dont,sotudeh-etal-2020-guir}, many models~\citep{wiedemann-etal-2020-uhh} achieved high performance on toxicity detection~\citep{zampieri-etal-2020-semeval}. However, directly deploying such systems could be problematic for the following reasons. Despite being highly effective, one major problem is that the decisions of the systems are largely opaque, i.e., it can be difficult to reason why the model made its decision~\citep{waseem-hovy-2016-hateful}. This interpretability problem is especially apparent when compared to prior models, such as Logistic Regression, that transparently assign scores to each input feature (here, words) that can be used to justify the model's decision. Knowing how one decision is made is important in toxicity detection. On one hand, recent laws such as General Data Protection Regulation\footnote{\url{https://gdpr-info.eu/}} highlighted the significance of interpretable models for users; on the other hand, interpretable models can assist online community moderators in reducing their time spent on checking each potentially problematic post. Since the purpose of these explanations is for human consumption, we consider a model to be \textit{interpretable} if it can produce a set of words from the input text that humans would consider a reasonable justification for the model's decision.

\begin{figure*}[ht!]
\centering
\begin{subfigure}[b]{0.48\linewidth}
\centering
\includegraphics[width=\textwidth]{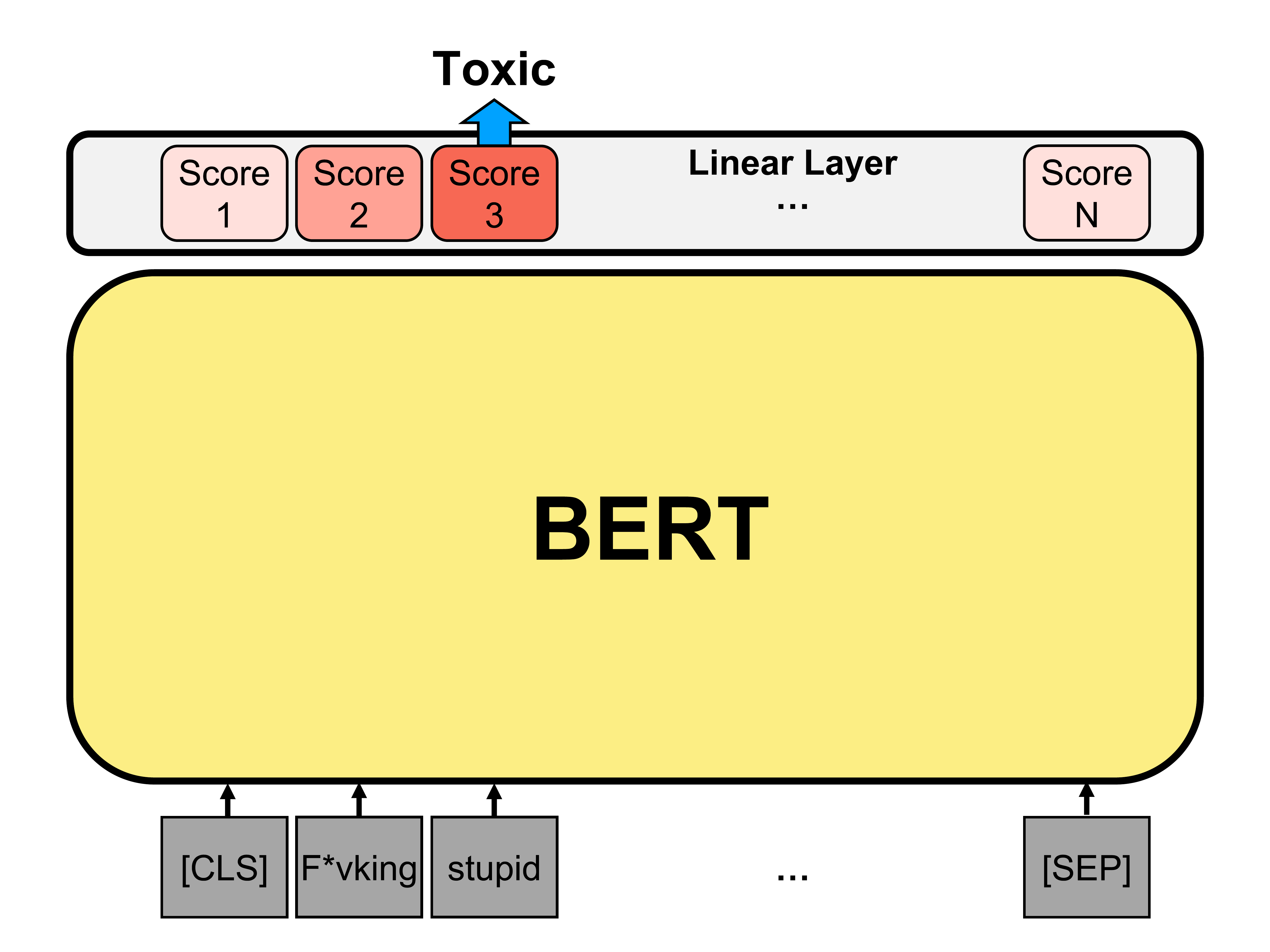}
\caption{When the input sequence is toxic, the toxicity of the most toxic span is picked to represent the toxicity of the sentence.}
\label{fig:model_illustration_t}
\end{subfigure}
\hfill
\begin{subfigure}[b]{0.48\linewidth}
\centering
\includegraphics[width=\textwidth]{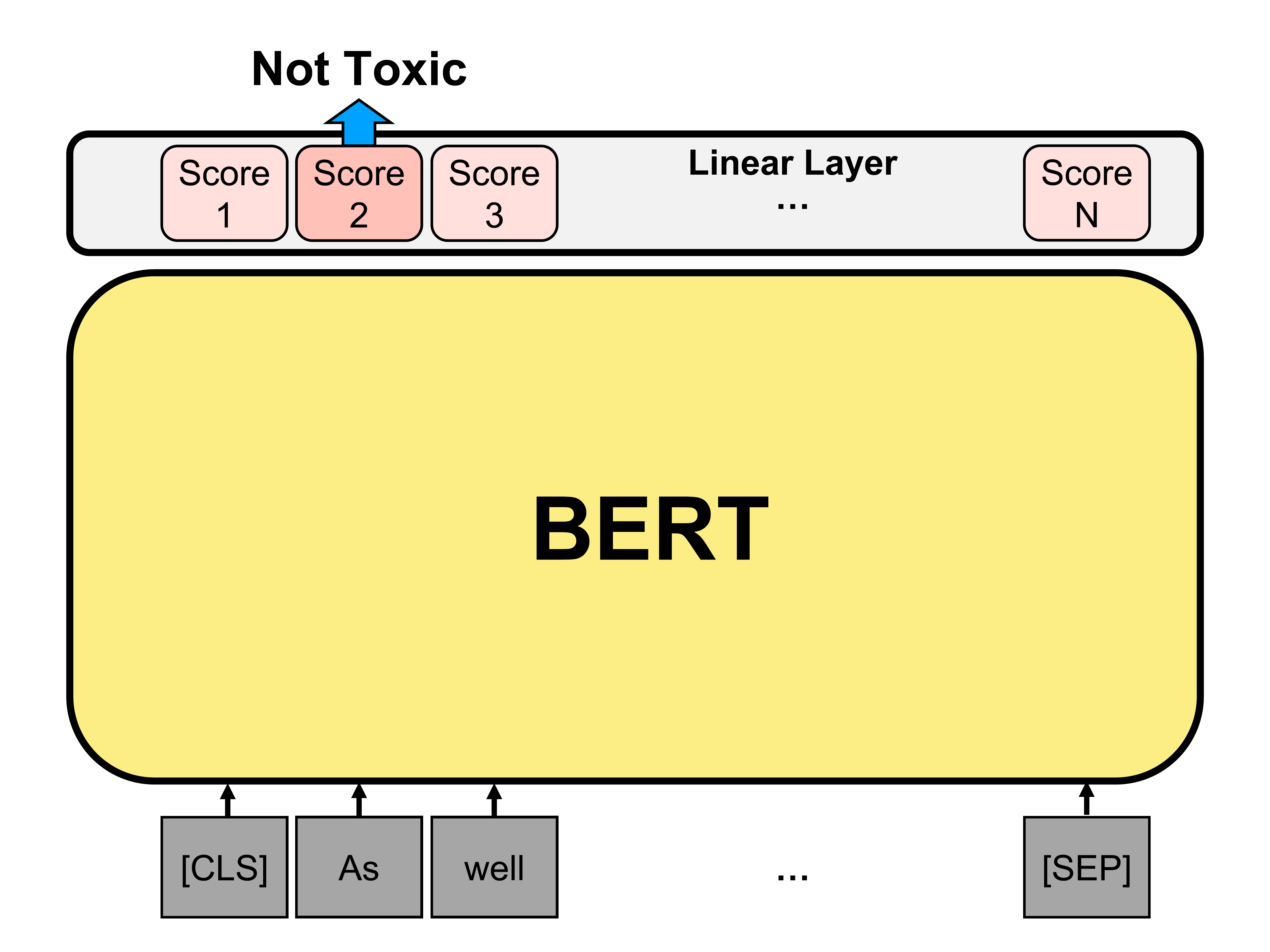}
\caption{When the input sequence is not toxic, none of the spans are toxic, and thus the whole sequence is predicted as non-toxic.}
\label{fig:model_illustration_nt}
\end{subfigure}
\caption{Illustration of our proposed approach. We showcase applying our proposed approach to the transformer-based model, BERT, in this case. In the linear layer, darker color denotes a more toxic span.}
\label{fig:model_illustration}
\end{figure*}

In this work, we propose a technique to improve the interpretability of transformer-based models like BERT~\citep{devlin-etal-2019-bert} and ELECTRA~\citep{Clark2020ELECTRA:} for the task of toxicity detection in social media posts. We base our technique on a simple and powerful assumption: \textit{A post is at least as toxic as its most toxic span}. In other words, the toxicity of a piece of text should be associated with the most toxic span identified in the text.
To this end, we propose using neural multi-task model that is trained on (1) toxicity detection over the entire piece of text and (2) toxic span detection (i.e., identifying individual tokens in the text that are toxic). Rather than the typical transformer classification approach, our model predicts the toxicity of each individual term in the text and aggregates them via max pooling to predict the toxicity of the entire text (see~\cref{fig:model_illustration}). Through experiments on the Civil Comment Dataset~\citep{10.1145/3308560.3317593}, we find our proposed approach not only improves the classification effectiveness compared to models that are only trained on the classification task, but also helps when transferring the model to a similar task. More importantly, however, the structure of our model inherently generates explanations of the decision by selecting the terms with the highest toxicity scores. We find through a human study that these explanations exceed the quality of those provided by Logistic Regression---a model often regarded as highly interpretable. 
An error analysis has shown multiple insights into utilizing contextualized models in toxicity detection and lead future directions.

\section{Related Work}
\label{sec:related}
Due to the ubiquity of online conversations, the need for automatic online toxicity detection has become crucial to promote healthy online discussions. Early research relied on surface-level features, e.g., bag-of-words approaches~\citep{warner-hirschberg-2012-detecting,waseem-hovy-2016-hateful,hateoffensive}, and traditional machine learning methods. Although reported to be highly predictive~\citep{schmidt-wiegand-2017-survey} and easily interpretable, they suffer from the problem of false positives as the presence of certain patterns could lead to misclassification~\citep{10.5555/2891460.2891697}. For example, some slurs that frequently appear in African American English are usually picked as strong evidence of toxicity~\citep{xia-etal-2020-demoting}; these words are indeed innocuous and only confined within the black online community. These features also require such predictive terms to appear in both training and testing set to work effectively. Later on, neural textual representations have shown effectiveness in toxicity detection. \Citet{10.1145/2740908.2742760} proposed using sentence-level embedding~\citep{10.5555/3044805.3045025} to represent the textual information and has shown great improvement against average over word-level embedding~\citep{10.1145/2872427.2883062}. These representations are usually utilized together with either linear classifiers such as Logistic Regression~\citep{10.1145/2740908.2742760}, or neural classifiers such as Convolutional Neural Networks (CNN)~\citep{gamback-sikdar-2017-using} and Long Short-Term Memory Networks (LSTM)~\citep{10.1145/3041021.3054223}. More recently, large-scale pre-trained language models such as Bidirectional Encoder Representation from Transformer (BERT)~\citep{devlin-etal-2019-bert} have shown great advantages in toxicity detection~\cite{zampieri-etal-2019-semeval, zampieri-etal-2020-semeval}, by learning contextual word embeddings instead of static embeddings.

Despite the fact that systems can achieve astonishing performance on given datasets, they suffer from the problem of lacking interpretability. One way of providing interpretability is to explain predictions~\cite{belinkov-glass-2019-analysis}. For textual data, such explanations could be those sub-strings (i.e., spans)
that significantly influence the models' judgments, which were named \textit{rationales} in~\citet{zaidan-etal-2007-using}. \Citet{zhang-etal-2016-rationale} proposed a CNN model that exploits document labels and associated rationales for text classification. On toxicity detection, while most of the works were done focusing on improving the model performances, less attention was paid to interpretability. SemEval-2021 Task 5 provided the Toxic Spans Detection Dataset (TSDD)\footnote{\url{https://sites.google.com/view/toxicspans}\label{tsdwebsite}} where each sample is annotated with both post-level label and rationales. \Citet{Mathew2020HateXplainAB} also provided a hate speech dataset where samples are annotated with rationales and other labels; though they also provided baseline models that can incorporate rationales, these models were built to evaluate the effectiveness of their proposed dataset, and thus cannot be easily transferred to other settings.

\section{Model}
\label{sec:model}
We propose a neural multi-task model that can predict the toxicity and explain its prediction at the same time by providing a set of words that can justify its prediction. In this section, we introduce the assumption that empowers our model with interpretability. Then we present the proposed model's architecture and the multi-task training paradigm.

\subsection{Assumption}
We begin with the following assumption: \textit{A post is at least as toxic as its most toxic span}. This assumption suggests that if there is a word or phrase in a piece of text that is toxic, i.e., with a level of toxicity that is over a certain threshold, the toxicity level of the entire text is certainly over such threshold, and, therefore, should be considered toxic.

This assumption can be formalized as follows. Let $\mathbf{x} = \{{x_1}, {x_2}, ..., {x_n}\}$ denote the input sequence where $n$ is the length of the sequence. Given the input $\mathbf{x}$, we can define $y$ as the toxicity label for the sequence and $\mathbf{y}$ as the toxicity labels for individual token. Let $\mathbf{s} = \{s_1, s_2, ..., s_n\}$ be a model's prediction of toxicity for each token.
By our assumption, we apply a max pooling operation over $\mathbf{s}$:
\begin{equation}
\label{eq:maxS}
\widetilde{s} = \max(\mathbf{s})
\end{equation}
where $\widetilde{s}$ represents the predicted toxicity of the entire sequence.

We acknowledge that this assumption may not always hold. In some cases, toxicity can be expressed in subtle or implicit ways, such as through sarcasm or metaphor~\citep{macavaney:plosone2019-hate,waseem-etal-2017-understanding}. In such cases, there is often not a clearly identifiable span that is toxic. However, these cases are difficult for any model to identify, and through our experimental results in~\cref{sec:results_toxic}, we find that this does not hinder the effectiveness of our model.

\subsection{Architecture \& Methodology}
\label{sec:arch}
To detect the toxicity and learn the toxic spans at the same time, we propose to use a neural multi-task learning framework~\citep{caruana1997multitask}. In our settings, we jointly train the model with two related tasks: (1) Toxicity Detection (at the sequence level), and (2) Toxic Span Detection (at the token level). These two tasks share all the parameters in the model. Our approach can be applied to any sequence encoder model (e.g., LSTM or transformer). Given the input sequence $\mathbf{x}$, let the output of the sequence encoder be $\mathbf{H} = \{\mathbf{h_1}, \mathbf{h_2}, ..., \mathbf{h_n}\}$, $\mathbf{H}\in\mathbb{R}^{n\times d}$. Here $\mathbf{h_i}\in\mathbb{R}^d$ denotes the $i$-th hidden state of the final layer, where $d$ denotes the length of the hidden embedding. Unlike what~\citet{devlin-etal-2019-bert} do for sequence classification where the linear layer is only stacked on the \texttt{[CLS]} token, we place it on top of the whole output sequence $\mathbf{H}$. Let $\mathbf{W}\in\mathbb{R}^{d \times 1}$ denote the parameters of the linear layer, then we have:
\begin{equation}
    \label{eq:S}
    \mathbf{s} = \mathbf{H}\cdot\mathbf{W}
\end{equation}
For the classification task, we use $\widetilde{s}$ as the predicted toxicity where $\widetilde{s}$ is calculated following the procedure mentioned in~\cref{eq:maxS}. For the span detection task, we directly leverage the output toxicity sequence $\mathbf{s}$. This setup ensures that the model learns to predict the text as toxic if a span is toxic.

For the purposes of training, let $\mathcal{D}_1$ be the dataset for toxicity detection task, and $\mathcal{D}_2$ be the dataset for toxic spans detection task. We construct the loss of the model $\mathcal{L}$ to be the following:
\begin{equation}
\centering
\begin{split}
    \label{eq:L}
    \mathcal{L} &= \lambda\underbrace{\sum_{(\mathbf{x}, y)\in\mathcal{D}_1}\mathcal{L}_{C}(\mathbf{x},y)}_\text{Loss for toxicity detection}\\
    &+ (1 - \lambda)\underbrace{\sum_{(\mathbf{x},\mathbf{y})\in\mathcal{D}_2}\mathcal{L}_{S}(\mathbf{x},\mathbf{y})}_\text{Loss for toxic spans detection}\\
\end{split}
\end{equation}
where $\mathcal{L}_{C}$ is the loss for the toxicity detection task and $\mathcal{L}_{S}$ is the loss for the toxic spans detection task. $\lambda$ denotes a hyperparameter specifying the weight for each task. Here the toxicity detection task is a sequence classification task and the toxic spans detection task is a token classification task. We jointly train the model across tasks in an end-to-end fashion, minimizing the Mean Square Error (MSE) loss for both tasks.

\begin{table*}[ht]
\centering
\begin{tabular}{l|c|c}
\toprule
\textbf{Comment}    & \textbf{Toxicity} & \textbf{Label}     \\ \hline
Like me flagging your comments.  & 0        & Non-toxic\\
We need to tax the clueless, irresponsible \underline{\colorbox{purple}{idiots}}. & 0.898 & Toxic\\
Don't take the bait of the \underline{\underline{\colorbox{pink}{troll}}}. It's what they want.   & 0.167    & Non-toxic \\
\bottomrule
\end{tabular}
\caption{Three samples from Curated CCD. We highlight the toxic spans in the toxic sample with \underline{\colorbox{purple}{purple underline}}. We mark the potentially toxic words in the non-toxic sample with \underline{\underline{\colorbox{pink}{pink double underline}}}.}
\label{tab:dataset}
\end{table*}

\section{Experiments}\label{sec:exp}

We conduct experiments to answer the following research questions:
\begin{enumerate}
\item[\textbf{RQ1}] Does our aforementioned assumption affect the model's performance at detecting toxic content?
\item[\textbf{RQ2}] Is our approach applicable to different transformer models?
\item[\textbf{RQ3}] Does our approach produce models that can generalize to different domains?
\item[\textbf{RQ4}] Does our model identify spans that improve the interpretability of model decisions?
\end{enumerate}

\subsection{Data}
\label{sec:data}
We primarily train and evaluate our system using the Civil Comment Dataset (CCD)~\citep{10.1145/3308560.3317593}. For interpretability, we leverage the Toxic Spans Detection Dataset (TSDD)\footref{tsdwebsite} %
in a multi-task training paradigm. 
We also use the Offensive Language Identification Dataset (OLID)~\cite{zampieri-etal-2019-predicting} for the cross-domain evaluation.

\noindent\textbf{CCD}\hspace{0.5cm}
The CCD~\citep{10.1145/3308560.3317593} is a large-scale dataset with crowd-sourced post-level annotations for toxicity, provided by the Civil Comment platform.\footnote{The platform was shut down by the end of 2017: \url{https://medium.com/@aja_15265/saying-goodbye-to-civil-comments-41859d3a2b1d}}
Posts that are rude, disrespectful, or unreasonable~\cite{10.1145/3308560.3317593} are considered toxic based on the rating guidelines as published by the Perspective API~\citep{Nithum2017WikipediaTL, 10.1145/3038912.3052591}. Through the platform, crowd-sourced raters were asked to rate comments as ``Very Toxic", ``Toxic", ``Hard to say", or ``Not Toxic". A toxicity score between zero and one of a post is the fraction of raters considering it to be toxic. We further cast the scores to binary labels by setting a threshold of 0.5 (i.e., at least half of the raters consider the post toxic). 
The dataset contains around 1.8 million posts in total, and 8\% of them are labeled as toxic.

\noindent\textbf{TSDD}\hspace{0.5cm}
TSDD\footref{tsdwebsite} is a 10,000-sample subset of CCD, containing only toxic comments, marked up with individual spans that are toxic. Each post is annotated by three annotators. 
528 posts have no annotated span since the annotators believe they are toxic as a whole without any explicit span. 

\noindent\textbf{OLID}\hspace{0.5cm}
We use the OLID offensive language dataset~\citep{zampieri-etal-2019-semeval} to examine the domain-transferability. This dataset contains web posts with hierarchical annotation~\citep{waseem-etal-2017-understanding}. We use its first layer, where the annotations indicate whether the content is offensive (32\%) or non-offensive (67\%). For our evaluation, we use the official 860-sample OLID test set.%

\noindent\textbf{Curated CCD}\hspace{0.5cm}
For the purpose of training and evaluation, we constructed a 30,000-sample curated CCD by mixing the clear-cut examples with the ambiguous ones, showcased in~\cref{tab:dataset}. Here, 14,000 samples are used for training and the rest are for testing.

We first sampled 7,000 highly toxic posts (toxicity score greater than 0.8) and 7,000 non-toxic posts (toxicity score less than 0.1) from CCD. Note that 3,000 of the toxic posts sampled were drawn from TSDD, which is still a subset of CDD, for the span annotations. These posts are considered to be \textit{easy} since a great portion of the raters agreed on the judgment. 

We further sampled another 8,000 ambiguous posts that have toxicity scores between 0.1 and 0.3 and contain terms that frequently appear in the toxic posts. Terms annotated at least 20 times as part of toxic spans in TSDD are considered frequent, resulting in a list of 62. The top 20 terms are presented in~\cref{tab:toxic word list}. We believe that these toxic terms are used in a non-toxic way in these posts, and, therefore, are good adversarial examples for the models to learn from the context instead of memorizing the frequent terms. To maintain an even proportion of toxic and non-toxic posts, we sample an additional 8,000 highly toxic posts.

\begin{table}[t]
\centering
\begin{tabular}{p{1.2cm}l|p{1.2cm}l}
\toprule
\textbf{Word} & \textbf{Freq.} $\downarrow$ & \textbf{Word} & \textbf{Freq.} $\downarrow$\\ \hline
stupid & 1085  & idiotic & 119 \\ 
idiot & 572  & ridiculous & 103 \\
idiots & 378  & ass & 102 \\
ignorant & 270  & fools & 100 \\
stupidity & 253  & damn & 100 \\
dumb & 185  & racist & 97 \\
moron & 163  & loser & 91 \\
fool & 163  & morons & 88 \\
pathetic & 145  & hypocrite & 65 \\
crap & 127  & shit & 62 \\
\bottomrule
\end{tabular}
\caption{The 20 most frequent entries of the toxic word list for finding ambiguous non-toxic samples in CCD.}
\label{tab:toxic word list}
\end{table}

\subsection{Implementation}
Here, we describe the implementation details of our baselines and the proposed models.

\noindent\textbf{LR}\hspace{0.5cm}
We use Logistic Regression (LR) classifier with lemmatized uni-gram and bi-gram features and L2 regularization as our baseline for both effectiveness evaluation and interpretability evaluation. LR is shown to be effective in toxicity detection~\citep{10.1145/2740908.2742760,waseem-hovy-2016-hateful} while naturally providing explanations for its predictions if utilized together with bag-of-tokens features. While there are other models that rely on the attention mechanisms to provide various interpretability~\citep{rogers-etal-2020-primer}, we choose to use LR instead of them. On the one hand, they usually need postprocessing to perform interpretation, which is less straightforward and intuitive compared to LR; on the other hand, there are still debates about whether attention mechanisms can produce meaningful explanations~\citep{wiegreffe-pinter-2019-attention, jain-wallace-2019-attention}. For our LR baseline, \texttt{NLTK} is used for preprocessing and \texttt{scikit-learn} is used for building the classification model.

For the following transformer-based models, we utilize the \texttt{transformers}~\citep{wolf-etal-2020-transformers} library with Adam~\citep{DBLP:journals/corr/KingmaB14} optimizer. The learning rate is by default set to $2\times10^{-5}$ and the number of training epochs is tuned on the validation set. We also utilized the pre-trained BERT-base~\citep{devlin-etal-2019-bert} and ELECTRA-base~\cite{Clark2020ELECTRA:} that are available in the Huggingface community.
Input sentences are trimmed to a max length of 256 tokens. 

\begin{table*}[t]
\centering
\begin{tabular}{l|ccc|ccc|c}
\toprule
 & \multicolumn{3}{c|}{\textbf{Non-Toxic}} & \multicolumn{3}{c|}{\textbf{Toxic}} &  \\
 \hline
\textbf{Model} & \textbf{P} & \textbf{R} & \textbf{F1} & \textbf{P} & \textbf{R} & \textbf{F1} & \textbf{Macro-F1} \\
\hline
LR & 0.824 & 0.828 & 0.826 & 0.827 & 0.823 & 0.825 & 0.826 \\
BERT-CLS & 0.944 & 0.674 & 0.786 & 0.746 & 0.960 & 0.840 & 0.813 \\
ELECTRA-CLS & 0.952 & 0.617 & 0.749 & 0.717 & 0.969 & 0.824 & 0.787 \\
BERT-SP & 0.843 & 0.771 & 0.806 & 0.789 & 0.857 & 0.822 & 0.814 \\
ELECTRA-SP & 0.884 & 0.716 & 0.791 & 0.761 & 0.906 & 0.827 & 0.809 \\
\hline
BERT-MT & 0.900 & 0.879 & 0.890 & 0.882 & 0.903 & 0.892 & 0.891 \\
ELECTRA-MT & 0.904 & 0.880 &\bf 0.892 & 0.883 & 0.907 &\bf0.894 &\bf 0.893 \\
 \bottomrule
\end{tabular}

\caption{In-domain evaluation results for toxicity classification on the Curated CCD dataset. We report Precision~(P), Recall (R), and F1 for each model on all categories. We also report the Macro-F1 for all models. The best F1 performance is indicated in \textbf{bold}.}
\label{tab:cls_eval}
\end{table*}

\noindent\textbf{BERT/ELECTRA-CLS}\hspace{0.5cm}
As baselines, we evaluate typical sentence classification (CLS) architectures tuned on only the post-level labels by adding a linear layer on top of the \texttt{[CLS]} token. Both BERT and ELECTRA are trained with the cross-entropy loss, which is the default setting in the \texttt{transformers} library. 

\noindent\textbf{BERT/ELECTRA-SP}\hspace{0.5cm}
We also evaluate models that are only trained for the span detection (SP) task. It follows the architecture and methodology mentioned in~\cref{sec:arch}, except only optimizing the toxic spans detection loss $\mathcal{L}_{S}$. Since only a portion of the samples contains toxic span annotations, the models are only trained on that subset. Beyond the toxic spans detection task, we also evaluate the toxicity detection performance for these models leveraging our proposed assumption, even though these models are not trained for toxicity detection.

\noindent\textbf{BERT/ELECTRA-MT}\hspace{0.5cm}
Finally, we describe the implementation of our proposed Multi-Task~(MT) models. 
The models are built with the architecture described in~\cref{sec:arch} and optimized with the joint loss $\mathcal{L}$ shown in~\cref{eq:L}. 
Since not all the input posts have the labels for toxic spans, the multi-task models are trained by interleaving samples with and without span information. Joint loss $\mathcal{L}$ is calculated for those samples with labels for both tasks; for samples with only post-level labels, we calculate only the classification loss $\mathcal{L}_{C}$. During training, we interleave the update with each kind of loss to ensure a balance update on the parameters. We specify the hyperparameter $\lambda$ to be 0.5 (i.e., weighting both tasks equally).

\section{Results on Toxicity Detection}\label{sec:results_toxic}
In this section, we answer the first three research questions by presenting and analyzing the classification effectiveness of the proposed models in various settings.

The in-domain toxicity detection performance is shown in~\cref{tab:cls_eval}. Both of our proposed multi-task models BERT-MT and ELECTRA-MT achieve the best performance among all models we evaluated. Models using the \texttt{[CLS]} tokens perform the worst among others, even worse than the ones that are only trained on the span detection task. This suggests that the span information is capable of predicting the toxicity of the entire content and combining it with the post-level supervised information further improves the effectiveness. This answers RQ1: our proposed model based on our assumption improves the classification effectiveness. Furthermore, BERT-MT and ELECTRA-MT are equally effective in respect to various evaluation metrics; therefore, we validate RQ2: our approach generalizes to at least two pre-trained transformer models.

Interestingly, LR has the highest precision among the baselines but the lowest recall, suggesting that the model may be relying on high-precision features such as racial slurs. This matches previous observations by \citet{macavaney:plosone2019-hate}.

In a domain transfer setting, i.e., training on CDD and testing on OLID, our approach is competitive with models trained on OLID. In~\cref{tab:cross_domain_eval}, both BERT-MT and ELECTRA-MT obtain a competitive 0.77 macro-F1 without training on any example in OLID, even outperforming some leading systems reported on this dataset. Therefore, we confirm that our models remain effective in a domain-transfer setting~(RQ3).

\begin{table}[t]
\centering
\begin{tabular}{l|c}
\toprule
\textbf{Model} & \textbf{Macro-F1} \\
\hline
SVM~\citep{zampieri-etal-2019-predicting}     &  0.69 \\
BiLSTM~\citep{zampieri-etal-2019-predicting}  &  0.75 \\
BERT-FT~\citep{liu-etal-2019-nuli}   & \textbf{0.83} \\  
\hline
BERT-MT (ours, transfer) &  0.77 \\
ELECTRA-MT (ours, transfer) & 0.77\\
\bottomrule
\end{tabular}
\caption{Evaluation on the OLID. BERT-FT stands for the BERT model fine-tuned on the OLID data. Our system performs competitive in a completely transfer setting (only trained on CCD data). We report Macro-F1 for all models here. The best performance is in \textbf{bold}.}
\label{tab:cross_domain_eval}
\end{table}
\subsection{Error Analysis}\label{sec:analysis}
To better understand the limitations of our proposed models, we qualitatively analyze the predictions against the gold labels using BERT-MT.

\noindent\textbf{False Positives}\hspace{0.5cm}
One major source of false positives comes from treating negative words, usually adjectives, such as \textit{disgusting}, \textit{lazy}, \textit{incompetent}, as strong signals for toxicity. 
Also, our proposed models tend to treat certain sub-words that frequently appear in toxic contexts as toxic spans. For example, the sub-word~\textit{\#\#nt} of \textit{magnificient} is predicted as toxic (0.979\footnote{Toxicity score.}) in the following sentence:

\begin{quote}
    \textit{Should have had the magnificient Doug Ford stump for Smith....LOL}
\end{quote}

This could be attributed to its high frequency in explicitly toxic words such as \textit{ignorant} and \textit{arrogant}.

We also suspect that the model is over-leveraging expression patterns or co-occurrences for toxicity classification. For example:

\begin{quote}
    \textit{I have never seen a suspect identified as a ``brown'' man.}
\end{quote}

Here, the model misclassifies it as toxic and picks the word \textit{brown} as the most toxic span with a toxicity score of 0.527. 
We first rule out the possibility that the toxicity comes from the negation by experimentally removing the word \textit{never} from the sentence. We find that the toxicity score of the sentence increases to 0.762 with the word \textit{brown} still as the most toxic span. We then examine by replacing the word \textit{suspect} and keep the rest; when we replace \textit{suspect} with \textit{girl} or \textit{mom}, the toxicity score for \textit{brown} decreases to 0.334 and 0.141 respectively; when changing to \textit{prisoner} or \textit{scammer}, the toxicity score for word \textit{brown} becomes 0.717 and 0.972. Among all the cases that we have experimented with, the word \textit{brown} is consistently the most toxic word in the sentence. It seems that the model is learning the correlation between the noun \textit{suspect} (or any other word positioned here) and the adjective \textit{brown}. Though it might lead to false positives in rare cases, the predictive power that enables this phenomenon could potentially be the reason that our model is doing better in picking unseen words as toxic spans than the LR model since it infers the prediction without leveraging on the lexical information but the syntactic information.

\noindent\textbf{False Negatives}\hspace{0.5cm}
Some comments are inherently hard for both human and machine classifiers to identify.
There are cases where the sentences can reasonably be considered as non-toxic where our model is also predicting as such. 
For example:
\begin{quote}
    \textit{Ignore the trolls Sheema......You are great and I always enjoy your pieces}.\\\\
    \textit{LOL....I keep hearing Garland never got a hearing blah,blah,blah....It sucks being in the minority....Go win some elections..}\\\\
    \textit{When the headline reads "Steve Bannon's porn and meth house"}
\end{quote}
In the first and second example, the speaker is not intended to be toxic even with the appearance of the word \textit{trolls} and \textit{sucks}. For the third example, the toxicity lies in the quoted text, which is also not the intention of the speaker. It is debatable that whether our model is indeed making mistakes on these naturally ambiguous comments.

Another major source of false negatives comes from the unawareness of the outside context, e.g., the target of the comment. For example:
\begin{quote}
    \textit{Degenerate comment.}\\\\
    \textit{I'm beginning to think the left lacks the mental capacity to reason.}
\end{quote}
These words/phrases such as \textit{degenerate}, and \textit{lacks the mental capacity to}, which could also be utilized in the neutral descriptions, are used here for expressing toxicity. It is easy for a human to reason out the target mentioned in the sentence and thus be aware of the toxicity raised; this is usually not the case for machine classifiers.

Also, many users intentionally modify the explicitly toxic words (e.g., replacing or removing characters) to obfuscate automatic detection while still keeping their intent clear to human~\citep{10.1145/2740908.2742760}, e.g., changing word \textit{idiots} into \textit{I.d.i.o.t.s} or modifying word \textit{asshole} to \textit{a-hole}. Even for advanced transformer models, they still need to learn deeper information or be incorporated with more human supervision to be fully aware of these cases.

\begin{table}[t]
    \centering
    \begin{tabular}{l|ccc}
    \toprule
        \textbf{Model}     & \textbf{SD-P}  & \textbf{SD-R} & \textbf{SD-F1}  \\ \hline
        LR    &  0.111  & 0.195  &  0.120\\
        BERT-SP    &  0.836 &  0.798 & {0.792} \\
        ELECTRA-SP & 0.840  &  0.807 & \textbf{0.798} \\
        \midrule
        BERT-MT    &  0.837 &  0.785 & 0.784 \\
        ELECTRA-MT & 0.842  & 0.788  & 0.789 \\
    \bottomrule
    \end{tabular}
    \caption{Evaluation results on toxic spans detection task. The best performance is \textbf{bolded}.}
    \label{tab:inter_eval}
\end{table}

\section{Interpretability}

Finally, we examine the interpretability of our models. We first leverage the existing span annotations to evaluate but discovered that they limit our analysis. To overcome the limitation, we further conduct a user study to evaluate the interpretability directly. 

\subsection{Span Detection as Interpretation}\label{sec:spandetection_result}
We select a balanced 8,000-sample set from the test split of the Curated CCD. Here, non-toxic posts have toxicities between 0 and 0.1 and therefore are all considered to have no toxic spans; toxic samples are all from TSDD.
We follow the ad-hoc evaluation metrics which are introduced in~\citet{da-san-martino-etal-2019-fine} and utilized in SemEval 2021 Task 5\footref{tsdwebsite} -- Span Detection Precision (SD-P), Recall (SD-R), and F1 (SD-F1). Due to the nature of the toxicity span detection task where instances span from single tokens to multiple sentences, the ad-hoc evaluation metrics give partial credits to imperfect matches at the character level. Given a post $t$, let the ground truth be a set of character offsets $S_{G}^{t}$ and let one certain system $A_i$ return a set of character offsets $S_{A_i}^{t}$. With the system $A_i$ and ground truth, the SD-P and SP-R on post $t$ are then defined as follow:
\begin{equation}
    \begin{split}
    P^t &= \frac{\left| S_{A_i}^{t} \cap S_{G}^{t} \right|}{\left| S_{A_i}^{t} \right|} \\
    R^t &= \frac{\left| S_{A_i}^{t} \cap S_{G}^{t} \right|}{\left| S_{G}^{t} \right|}
    \end{split}
\end{equation}
With SD-P and SD-R defined, the SD-F1 is also defined:
\begin{equation}
    F_1^t = \frac{2 \cdot P^t \cdot R^t}{P^t + R^t}
\end{equation}

We report the average values over all samples. The evaluation results are shown in~\cref{tab:inter_eval}. 

Models trained for the span detection task (*-SP) achieve the highest F1 scores with no surprise. However, our multi-task models provide nearly equal span detection effectiveness with far better performance on toxicity detection (See~\cref{tab:cls_eval}). 
Also, BERT-MT and ELECTRA-MT perform similarly here, further confirming that our approach can be used in various transformer models (RQ2).

We also compared with LR, which is widely considered to be interpretable. The predicted span from the LR model is reconstructed from the bag-of-token features. Features that contribute to the positive score for the sample are considered as the \textit{interpretation} of the model. Our proposed models were shown to strongly outperform LR by a large margin. Note that LR's performance is hindered in this evaluation, as it was not trained for sequence classification (e.g., what CRF would do), despite being considered an interpretable model.

\subsection{User Study}

The labeled toxic spans used for the evaluation in~\cref{sec:spandetection_result} are not annotated to be the \textit{interpretation} for the toxicity of the post. We argue that it is not sufficient to evaluate the interpretability of prediction along with other known issues in automated evaluation~\citep{manning-etal-2020-human,novikova-etal-2017-need}.

Therefore, we select 400 toxic samples from the test split of Curated CCD and 237 offensive samples from the testing set of OLID for the interpretability study.
For each sample, three words with the highest predicted score from the models are picked as the explanation and are post-processed to the same form to avoid identifying the models based on the types of token preprocessing. 
Annotators are asked to annotate for the toxicity (e.g., whether the sample is toxic) of the samples and pick the model with a better explanation. 
The order and the name of the models are hidden from the annotators to avoid biases. 
Although ELECTRA-MT and BERT-MT perform comparably in terms of F1, we found that qualitatively BERT-MT is better and therefore picked for the user study.

Despite the subjectivity of the annotation task, the annotators agreed in 78\% of the cases. For our quantitative analysis, we use the samples that the annotators agree on. We also filter out 22 samples that both of the annotators considered non-toxic.

\begin{table}[h]
\centering
\begin{tabular}{l|c|l|c|c}
\toprule
\textbf{Preference} & \multicolumn{2}{c|}{\textbf{CCD}}  & \textbf{OLID}   & \textbf{All}    \\
\hline
LR               & \multicolumn{2}{c|}{26.5\%} & 18.8\%   & 24.0\%   \\
BERT-MT             & \multicolumn{2}{c|}{59.4\%} & 41.6\%   & 53.7\%   \\
No Preference    & \multicolumn{2}{c|}{14.2\%} & 39.6\%   & 22.3\%   \\
\bottomrule
\end{tabular}
\caption{Aggregated human preference during the interpretability experiment for CCD and OLID. We see that the annotators prefer the explanations from the BERT-MT over those from LR by a hefty margin.}
\label{tab:int_annotation}
\end{table}

On average, our annotators prefer BERT-MT over LR on more than half of the samples. As shown in~\cref{tab:int_annotation}, BERT-MT is considered to be more interpretable in both datasets by a wide margin. This result not only suggests that our proposed assumption provides interpretability to transformer models (RQ4) but also better explanations than the widely-known interpretable models. 

\begin{table*}[t]
    \centering
    \small
    \begin{tabular}{c|c|l}
    \toprule
        \textbf{\#} & \textbf{Model} & \textbf{Example}\\
        \hline
        \\[-0.9em]
        \multirow{4}{*}{{A}} & {BERT-MT} & 
\colorbox{lightred!58.62501859664917}{\strut red}
\colorbox{lightred!98.29542636871338}{\strut \#\#neck}
\colorbox{lightgreen!12.507778406143188}{\strut followers}
\colorbox{lightgreen!77.16631293296814}{\strut are}
\colorbox{lightgreen!87.31656074523926}{\strut doing}
\colorbox{lightgreen!91.95192158222198}{\strut all}
\colorbox{lightgreen!80.52753806114197}{\strut they}
\colorbox{lightgreen!87.75179907679558}{\strut can}
\colorbox{lightgreen!84.06547755002975}{\strut to}
\colorbox{lightgreen!79.88553643226624}{\strut make}
\colorbox{lightgreen!73.20240437984467}{\strut america}
\colorbox{lightgreen!24.186033010482788}{\strut white}
\colorbox{lightgreen!74.39287006855011}{\strut again}
\colorbox{lightgreen!97.62225970625877}{\strut .}
 \\
        \\[-0.9em]
        \cline{2-3} 
        \\[-0.9em]
         & {LR} & 
\colorbox{lightred!0.0}{\strut redneck}
\colorbox{lightred!15.430800145100276}{\strut follower}
\colorbox{lightred!0.0}{\strut are}
\colorbox{lightgreen!11.891768224266086}{\strut doing}
\colorbox{lightred!0.0}{\strut all}
\colorbox{lightred!0.0}{\strut they}
\colorbox{lightred!0.0}{\strut can}
\colorbox{lightred!0.0}{\strut to}
\colorbox{lightred!0.5804311505935189}{\strut make}
\colorbox{lightred!0.5804311505935189}{\strut america}
\colorbox{lightred!31.85298126760385}{\strut white}
\colorbox{lightred!0.0}{\strut again}
\colorbox{lightred!0.0}{\strut .} \\
         \\[-1em]
        \hline
        \\[-0.9em]
        \multirow{4}{*}{{B}} & {BERT-MT} & \colorbox{lightgreen!94.81832943856716}{\strut not}
\colorbox{lightgreen!89.24821466207504}{\strut only}
\colorbox{lightgreen!84.32392925024033}{\strut is}
\colorbox{lightgreen!84.47504341602325}{\strut he}
\colorbox{lightgreen!76.11839175224304}{\strut a}
\colorbox{lightred!100.92060565948486}{\strut liar}
\colorbox{lightgreen!79.19858992099762}{\strut ,}
\colorbox{lightgreen!77.725949883461}{\strut he}
\colorbox{lightgreen!74.16299879550934}{\strut ’}
\colorbox{lightgreen!83.61876010894775}{\strut s}
\colorbox{lightgreen!78.13908457756042}{\strut a}
\colorbox{lightred!81.0981035232544}{\strut coward}
\colorbox{lightgreen!81.7324846982956}{\strut too}
\colorbox{lightgreen!66.46856665611267}{\strut !}
\colorbox{lightgreen!89.2903558909893}{\strut the}
\colorbox{lightgreen!27.24052667617798}{\strut clown}
\colorbox{lightgreen!89.38200175762177}{\strut is}
\colorbox{lightgreen!87.73523196578026}{\strut going}
\colorbox{lightgreen!93.10848489403725}{\strut down}
\colorbox{lightgreen!92.80130863189697}{\strut !} \\
        \\[-0.9em]
        \cline{2-3} 
        \\[-0.9em]
         & {LR} & 
\colorbox{lightred!0.0}{\strut not}
\colorbox{lightred!0.0}{\strut only}
\colorbox{lightred!0.0}{\strut is}
\colorbox{lightred!0.0}{\strut he}
\colorbox{lightred!0.0}{\strut a}
\colorbox{lightred!90.39167290828003}{\strut liar}
\colorbox{lightred!0.0}{\strut ,}
\colorbox{lightred!0.0}{\strut he}
\colorbox{lightred!0.0}{\strut ’}
\colorbox{lightred!0.0}{\strut s}
\colorbox{lightred!0.0}{\strut a}
\colorbox{lightred!67.1350349739778}{\strut coward}
\colorbox{lightred!0.0}{\strut too}
\colorbox{lightred!0.0}{\strut !}
\colorbox{lightred!0.0}{\strut the}
\colorbox{lightred!81.29625048215058}{\strut clown}
\colorbox{lightred!0.0}{\strut is}
\colorbox{lightgreen!9.11436158269041}{\strut going}
\colorbox{lightred!0.0}{\strut down}
\colorbox{lightred!0.0}{\strut !}\\
         \\[-1em]
         \hline
         \\[-0.9em]
         \multirow{4}{*}{{C}} & {BERT-MT} & 
\colorbox{lightgreen!103.27119454741478}{\strut wow}
\colorbox{lightgreen!99.66760203242302}{\strut you}
\colorbox{lightgreen!83.85075628757477}{\strut ’}
\colorbox{lightgreen!94.04607824981213}{\strut re}
\colorbox{lightgreen!63.836079835891724}{\strut un}
\colorbox{lightgreen!78.13134789466858}{\strut -}
\colorbox{lightgreen!61.65475249290466}{\strut smart}
\colorbox{lightgreen!104.94985729455948}{\strut .}
 \\
         \\[-0.9em]
         \cline{2-3}
         \\[-0.9em]
         & {LR} & 
\colorbox{lightred!7.199421911750292}{\strut wow}
\colorbox{lightred!0.0}{\strut you}
\colorbox{lightred!0.0}{\strut ’}
\colorbox{lightred!0.0}{\strut re}
\colorbox{lightred!0.0}{\strut un-smart}
\colorbox{lightred!0.0}{\strut .}\\
         \\[-1em]
         \hline
         \\[-0.9em]
         \multirow{4}{*}{{D}} & {BERT-MT} & \colorbox{lightgreen!88.47437128424644}{\strut your}
\colorbox{lightgreen!78.71609926223755}{\strut constant}
\colorbox{lightgreen!73.5869824886322}{\strut ability}
\colorbox{lightgreen!81.14759922027588}{\strut at}
\colorbox{lightgreen!75.28499364852905}{\strut being}
\colorbox{lightgreen!69.94664669036865}{\strut an}
\colorbox{lightgreen!9.840011596679688}{\strut a}
\colorbox{lightgreen!11.375504732131958}{\strut -}
\colorbox{lightgreen!7.8248679637908936}{\strut hole}
\colorbox{lightred!101.03647708892822}{\strut liar}
\colorbox{lightgreen!93.08888912200928}{\strut doesn}
\colorbox{lightgreen!90.5108779668808}{\strut '}
\colorbox{lightgreen!93.18382143974304}{\strut t}
\colorbox{lightgreen!92.59714633226395}{\strut make}
\colorbox{lightgreen!92.856365442276}{\strut you}
\colorbox{lightgreen!90.32100662589073}{\strut great} \\
         \\[-1em]
         \cline{2-3}
         \\[-0.9em]
         & {LR} & 
\colorbox{lightred!0.0}{\strut your}
\colorbox{lightgreen!11.513825497886355}{\strut constant}
\colorbox{lightgreen!30.271716353656142}{\strut ability}
\colorbox{lightred!0.0}{\strut at}
\colorbox{lightred!0.0}{\strut being}
\colorbox{lightred!0.0}{\strut an}
\colorbox{lightred!0.0}{\strut a-hole}
\colorbox{lightred!90.39167290828003}{\strut liar}
\colorbox{lightgreen!6.532196599128726}{\strut doe}
\colorbox{lightred!0.0}{\strut n't}
\colorbox{lightgreen!6.01136912678899}{\strut make}
\colorbox{lightred!0.0}{\strut you}
\colorbox{lightgreen!7.521704837327537}{\strut great}\\
    \bottomrule
    \end{tabular}
    \caption{Examples for qualitative analysis on interpretability. Here \colorbox{lightred!80.0}{\strut red color} indicates toxicity, \colorbox{lightgreen!80}{\strut blue color} indicates innocuousness, and white color represents neutral. Darker colors indicate more polarity.}
    \label{tab:examples}
\end{table*}
We now take a closer look at how BERT-MT model's explanations compare to those from LR. Some examples are shown in~\cref{tab:examples}. We find that the predictions from BERT-MT are more polarized, while those from LR tend to be neutral. This is because the usage of MSE loss in BERT-MT penalizes values for not being close to 0 or 1; for LR, the L2 regularization penalizes large weight values. It is also worth noting that these models differ in how term scores are aggregated over the post; BERT-MT takes the maximum, whereas LR takes the sum. For analysis, we highlight three cases here:

\noindent\textbf{When BERT-MT is preferred.}\hspace{0.5cm}
BERT-MT can pick out the toxic spans more accurately by leveraging the context. In comparison, the LR model suffers from picking words based on frequency even with a non-toxic usage, resulting in a bias toward some entities, such as certain groups of people (See example A in~\cref{tab:examples}).
BERT-MT also takes advantage of the WordPiece tokenization which retains the toxic word pieces when users are altering words to avoid censorship; while LR lemmatizes the tokens and resulting in losing this information. 

\noindent\textbf{When LR is preferred.}\hspace{0.5cm}
We observe that LR tends to predict more spans than BERT-MT, leading to higher recall, resulting in more \textit{comprehensive} explanations. Take the example B in~\cref{tab:examples} for instance, the LR is able to pick out the word \textit{coward} and \textit{clown} which are considered to be non-toxic by BERT-MT. We conclude that BERT-MT is more cautious in predicting words as toxic.

\noindent\textbf{When both models are equal.}\hspace{0.5cm}
We find that both models are doing poorly if the sentence is implicitly toxic. Such indirect toxicity is carried either by negation (Example C in~\cref{tab:examples}) or adversarially-modified toxic words (Example D in~\cref{tab:examples}).
Finally, there are also many cases where both models are doing equally well. This category of samples is generally easier to detect with explicit terms or slurs without ambiguity. 

\section{Conclusion \& Future Work}
In this paper, we proposed a toxicity detection approach that builds in the interpretability by predicting the toxicity of a piece of text based on the toxicity level of its spans. We showed that our approach is more effective in both in-domain and cross-domain evaluation than baselines that were shown to be effective. By conducting a user study, we further showed that our approach generates better explanations of the classification decisions than what Logistic Regression produces, which is known to be interpretable. 

In the future, we plan to extend our current work in several ways. We plan to take the implicit toxicity into consideration to make our assumption more robust. Besides, we will dig more into toxicity detection with long sequences. We also plan to investigate methods that consider more subtle context and actors in the content to better distinguish different usages of the toxic terms. 

\section*{Acknowledgements}

This work was supported in part by the ARCS Foundation.

\bibliography{eacl2021}
\bibliographystyle{acl_natbib}

\end{document}